\documentclass[runningheads]{llncs}
\usepackage[T1]{fontenc}
\usepackage{graphicx}
\usepackage{amsmath}
\usepackage{graphicx}
\usepackage{wasysym}
\usepackage{hyperref}
\usepackage{adjustbox}
\usepackage{booktabs}
\usepackage{multirow}
\usepackage{array}      
\usepackage{tikz}

\usepackage{xcolor}
\usepackage{lipsum}
\begin{document}
\title{Multi-Centre Validation of a Deep Learning Model for Scoliosis Assessment}
%
%
\author{Šimon Kubov\inst{1} \and
Simon Klíčník\inst{2} \and
Jakub Dandár\inst{3} \and
Zdeněk Straka\inst{3}\orcidID{0000-0002-2788-1667} \and
Karolína Kvaková\inst{3} \and
Daniel Kvak\inst{3,4}\orcidID{0000-0001-7808-7773}
}
\authorrunning{Š. Kubov et al.}
%
\institute{Department of Radiogiagnostics, Thomayer University Hospital, Prague, Czechia \and
Department of Radiology and Nuclear Medicine, University Hospital Královské Vinohrady and 3rd Faculty of Medicine, Prague, Czechia \and
Research \& Development Department, Carebot, Ltd., Prague, Czechia \\
\email{daniel.kvak@carebot.com}\\
\and
Department of Simulation Medicine, Masaryk University, 625 00 Brno, Czechia}
\maketitle              
\begin{abstract}
Scoliosis affects roughly 2--4\,\% of adolescents, and treatment decisions depend on precise Cobb‑angle measurement. Manual assessment is slow and subject to inter‑observer variation. We conducted a retrospective, multi‑centre evaluation of a fully automated deep‑learning software (Carebot AI Bones, Spine Measurement functionality; Carebot s.r.o.) on 103 standing anteroposterior whole‑spine radiographs collected from ten hospitals. Two musculoskeletal radiologists independently measured each study and served as reference readers. Agreement between the AI and each radiologist was assessed with Bland--Altman analysis, mean absolute error (MAE), root-mean-squared error (RMSE), Pearson correlation coefficient, and Cohen's~$\kappa$ for four‑grade severity classification. Against Radiologist~1 the AI achieved an MAE of $3.89^{\circ}$ (RMSE $4.77^{\circ}$) with a bias of $0.70^{\circ}$ (limits of agreement $-8.59^{\circ}$ to $9.99^{\circ}$); against Radiologist~2 it achieved an MAE of $3.90^{\circ}$ (RMSE $5.68^{\circ}$) with a bias of $2.14^{\circ}$ (limits $-8.23^{\circ}$ to $12.50^{\circ}$). Pearson correlations were $r = 0.906$ and $r = 0.880$ (inter‑reader $r = 0.928$), while Cohen's~$\kappa$ for severity grading reached 0.51 and 0.64 (inter‑reader $\kappa = 0.59$). These results show that the proposed software reproduces expert‑level Cobb‑angle measurements and categorical grading across multiple centres, suggesting its utility for streamlining scoliosis reporting and triage in clinical workflows.

\keywords{Artificial intelligence \and Cobb angle \and Deep learning \and Interobserver agreement \and Scoliosis \and Spine deformities}

\end{abstract}

\section{Introduction}

Scoliosis, a three-dimensional deformity characterised radiographically by a lateral curvature of the spine, affects roughly 2–4\% of adolescents and up to 8\% of adults worldwide \cite{konieczny2013epidemiology}. The magnitude of curvature, expressed by the Cobb angle on an upright anteroposterior (AP) full-length spine X-ray, determines both the diagnosis ($\geq$ 10°) and the clinical pathway: curves of 10–24° are usually observed, 25–39° prompt bracing, and angles $\geq$ 40–50° often lead to surgical referral \cite{weinstein2013effects}. Precise measurement is therefore critical during screening and longitudinal follow-up, yet manual Cobb angle tracing is time-consuming and suffers from inter-observer discrepancies of 5–10° even among experienced readers \cite{gstoettner2007inter}.


Over the last years, deep‐learning approaches have shown considerable promise in automating Cobb‑angle assessment. Convolutional neural‑network pipelines that first localise vertebral landmarks and then derive the maximal end‑plate angle have demonstrated close agreement with expert measurements in single‑centre studies \cite{horng2019cobb,sun2022comparison}. Building on these results, Zhu et al. conducted a comprehensive meta‑analysis of 14 models and confirmed that segmentation‑based architectures generally outperform landmark‑based ones \cite{zhu2025deep}. More recently, multi‑centre evaluations encompassing both adult and paediatric radiographs have reported near‑radiologist reliability across different vendors and demographic groups \cite{hayashi2024deep,wong2023validation}. Despite this progress, most published algorithms have been developed and tested on retrospective datasets from single institutions or on narrowly defined populations (e.g.\ adolescent idiopathic scoliosis), which limits their generalisability. 

To address this gap, we developed a deep‑learning model that (i) detects vertebral landmarks on whole‑spine anteroposterior (AP) radiographs and (ii) calculates the maximal Cobb angle to categorise scoliosis severity.  
In this multi‑centre validation study, we test the hypothesis that the model’s severity classification achieves agreement comparable to expert musculoskeletal radiologists when benchmarked against reference Cobb angles measured independently by two such readers.



\section{Materials and Methods}

\subsection{Software}

\begin{figure}[!ht]
\centering
\includegraphics[width=1\textwidth]{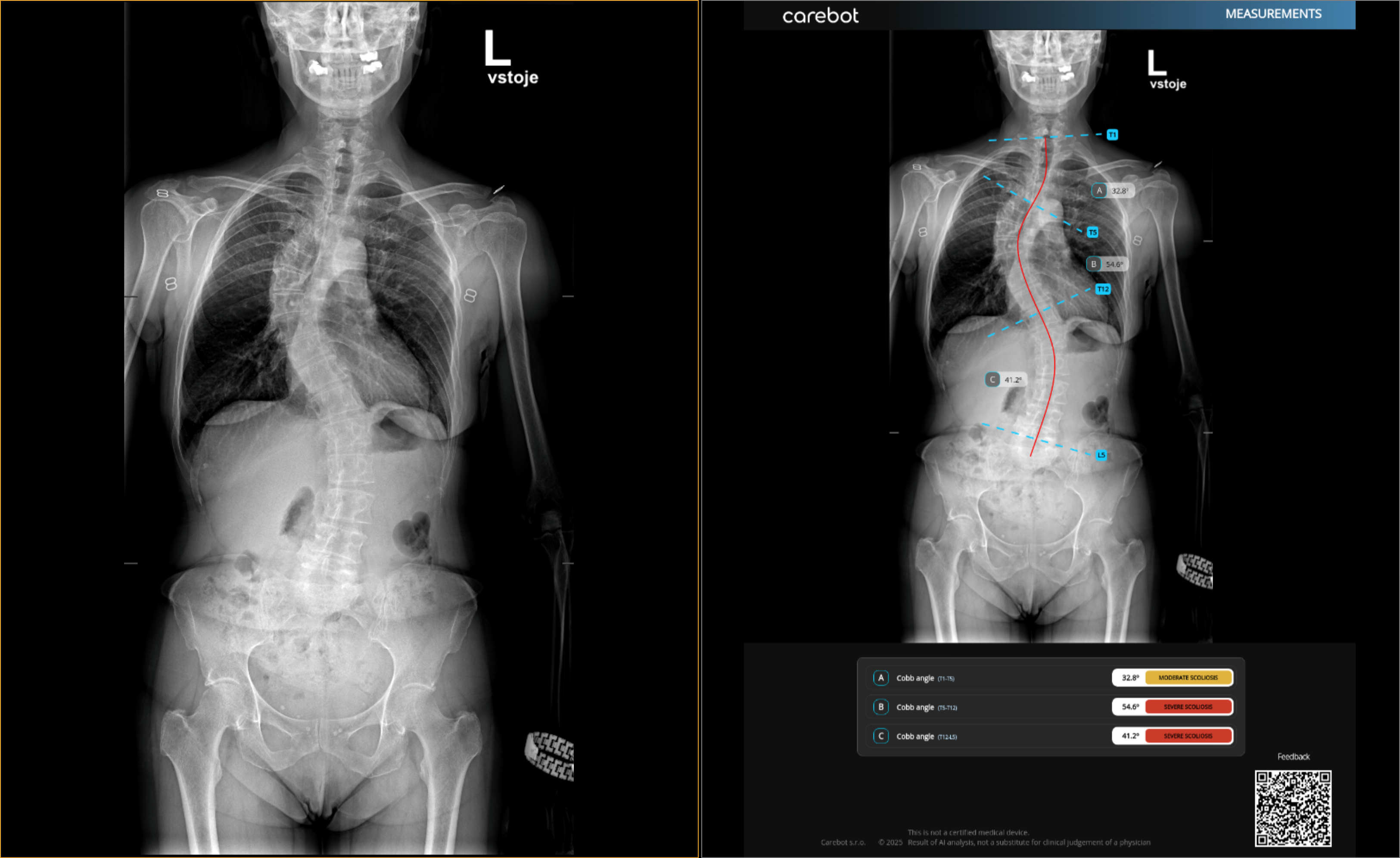}
\caption{\label{fig:demo}Demo.}
\end{figure}

The proposed AI software is a deep-learning-based solution (Carebot AI Bones, Spine Measurement functionality; Carebot s.r.o., Czechia) that detects vertebral landmarks on whole-spine anterior-posterior (AP) X-rays, and computes the Cobb angle to classify severity of the scoliosis as no scoliosis (< 10°), mild (10–24°), moderate (25–39°) or severe ($\geq$ 40°). The software implements a two-stage approach: (a) YOLOv11 landmark detector, trained on 575 expertly annotated whole-spine X-ray images (i.e. 39,100 vertebral keypoints) to localise the superior and inferior corners of vertebrae C7–L5; and (b) geometry‑based algorithm computes Cobb angles from these landmarks. Delivered as a self-contained application, it integrates directly into clinical PACS via both DICOMweb and DIMSE (\autoref{fig:demo}), automating study retrieval and result insertion. 

\newpage

\subsection{Data collection}


A total of 125 full-length standing anterior-posterior spine X-rays acquired between 1 and 18 May 2025 were collected from 10 participating hospitals. After applying inclusion and exclusion criteria (\autoref{tab:eligibility}), 22 studies were excluded and the final analysis set comprised 103 radiographs collected across ten hospitals—Hospital Hořovice ($n=48$), University Hospital Olomouc ($n=26$), Hospital Frýdek-Místek ($n=12$), Hospital Karviná-Ráj ($n=4$), Surgical Disciplines Prague ($n=3$), Hospital AGEL Nový Jičín ($n=3$), Třinec Hospital ($n=2$), Silesian Hospital in Opava ($n=2$), Regional Hospital Náchod ($n=2$), and Regional Hospital Příbram ($n=1$). Given its retrospective nature and full anonymization, the requirement of informed consent was waived. All DICOM files were deidentified according to PS 3.15 Basic Application Confidentiality Profile, with removal of direct identifiers.

\begin{table}[ht]
  \centering
  \caption{Eligibility inclusion and exclusion criteria.}
  \label{tab:eligibility}
  \begin{tabular}{@{}lp{9cm}@{}}
    \toprule
    \textbf{Criterion} & \textbf{Detail} \\ 
    \midrule
    Inclusion &
      All standing AP spine radiographs acquired between 1 and 18 May 2025 from patients aged $\geq$ 1 year. \\[4pt]
    Exclusion &
      (i) unreadable or incomplete images; 
      (ii) postoperative instrumentation obscuring $\geq$ 3 vertebrae; 
      (iii) duplicate studies. \\
    \bottomrule
  \end{tabular}
\end{table}

\newpage

\begin{figure}[!ht]
\centering
\includegraphics[width=0.9\textwidth]{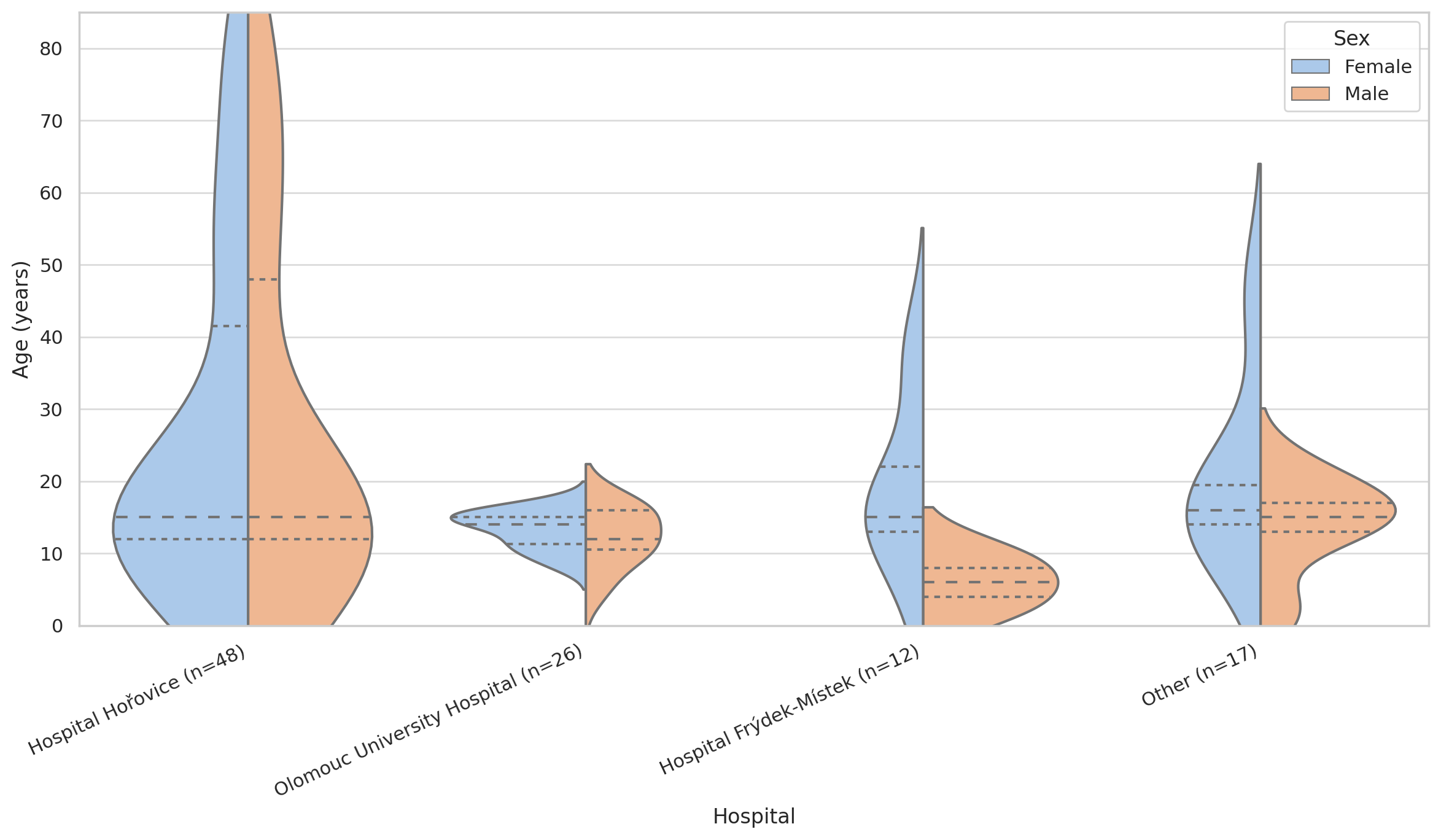}
\caption{\label{fig:demographics}Age distribution of patients by hospital and sex. The violin plot displays kernel density estimation of age for each group, split by sex (Female = blue, Male = orange).}
\end{figure}

The cohort demonstrated a predominantly paediatric and adolescent profile: mean age 18.6 ± 13.3 years, median 14 years (interquartile range 11–16 years), range 1.8–62.5 years; 81\% of radiographs were from patients younger than 20 years (\autoref{fig:demographics}). Females slightly outnumbered males, accounting for 60 (58\%) versus 43 (42\%) X-rays, respectively.

\subsection{Study Design}

This retrospective, multi-centre diagnostic accuracy study included all standing AP full-spine X-rays. Two musculoskeletal radiologists (Radiologist 1, Radiologist 2) independently measured the maximal Cobb angle on each X-ray using the institutional PACS tool. In parallel, the AI software analysed the same DICOM images, without manual intervention, to yield one automated Cobb angle per study. Readers and AI were blinded to each other’s results during measurement. For analysis, each AI measurement was compared to RAD1 and to RAD2 using paired design.

\subsection{Statistical Analysis and Endpoints}

The agreement of continuous measurements was evaluated by Bland-Altman analysis. For each subject \(i\), the difference \(d_i = m_{1,i} - m_{2,i}\) was computed (where \(m_{1,i}\) and \(m_{2,i}\) are two readers’ angles), and the bias \(\bar{d} = \tfrac{1}{N}\sum_i d_i\) was reported along with limits of agreement \(\bar{d}\pm1.96\,s_d\) (with \(s_d\) the SD of \(d_i\)). Mean absolute error (MAE = \(\tfrac{1}{N}\sum_i|d_i|\)) and root-mean-squared error (RMSE = \(\sqrt{\tfrac{1}{N}\sum_i d_i^2}\)) were also calculated. Continuous Cobb angles from the AI software and each radiologist were reported as mean ± standard deviation (SD). The linear association between two readers was quantified by Pearson’s correlation coefficient,  
\[
r = \frac{\sum_i(x_i - \bar{x})(y_i - \bar{y})}{\sqrt{\sum_i(x_i - \bar{x})^2\,\sum_i(y_i - \bar{y})^2}}\,,
\]  
where \(x_i\) and \(y_i\) are paired measurements and \(\bar{x},\bar{y}\) their means. For categorical agreement, Cobb angles were stratified into four grades (no scoliosis < 10°, mild 10–24°, moderate 25–44°, severe $\geq$ 45°). Pairwise agreement was summarized by Cohen’s kappa,  
\[
  \kappa = \frac{p_o - p_e}{1 - p_e}
\]

where \(p_o\) is the observed agreement and \(p_e\) the chance agreement. Before data extraction, a power analysis determined the minimum sample size required for achievement of endpoints (\autoref{tab:endpoints}). Assuming a standard deviation of absolute errors
\(\sigma=4.5^{\circ}\),
at least 
\[
n = \left( \frac{1.96 \times 4.5}{1.0} \right)^{2} \approx 78
\]

whole-spine X-rays were needed to bound the two‑sided 95 \% CI around the sample MAE within \(\pm1^{\circ}\). For the secondary endpoint, an expected \(\kappa=0.60\) and class distribution \(\{60\%,25\%,10\%,5\%\}\) (chance agreement \(P_{e}=0.435\)) yielded \(n\approx94\) images to keep the 95 \% CI of \(\kappa\) within \(\pm0.15\) (Sim–Wright approximation); a conservative Donner–Eliasziw test for demonstrating \(\kappa>0.40\) with 80 \% power suggested \(n\approx108\). All analyzes were conducted in Python (v3.10) using NumPy, pandas and scikit-learn.

\begin{table}[h]
\centering
\caption{Pre‐specified endpoints and corresponding metrics.}
\label{tab:endpoints}
\begin{tabular}{ll}
\toprule
Endpoint          & Metric \\ 
\midrule
Primary           & Bland–Altman analysis (bias and limits of agreement, bias ± 1.96 SD) \\ 
                  & Mean absolute error (MAE) and root mean squared error (RMSE) \\
\addlinespace
Secondary         & Mean ± SD of Cobb angles (°) \\
                  & Pearson’s \(r\) between paired measurements \\
                  & Cohen’s \(\kappa\) for four-grade severity \\

\bottomrule
\end{tabular}
\end{table}


\newpage

\section{Results}

\begin{figure}[!ht]
  \centering
  \includegraphics[width=\textwidth]{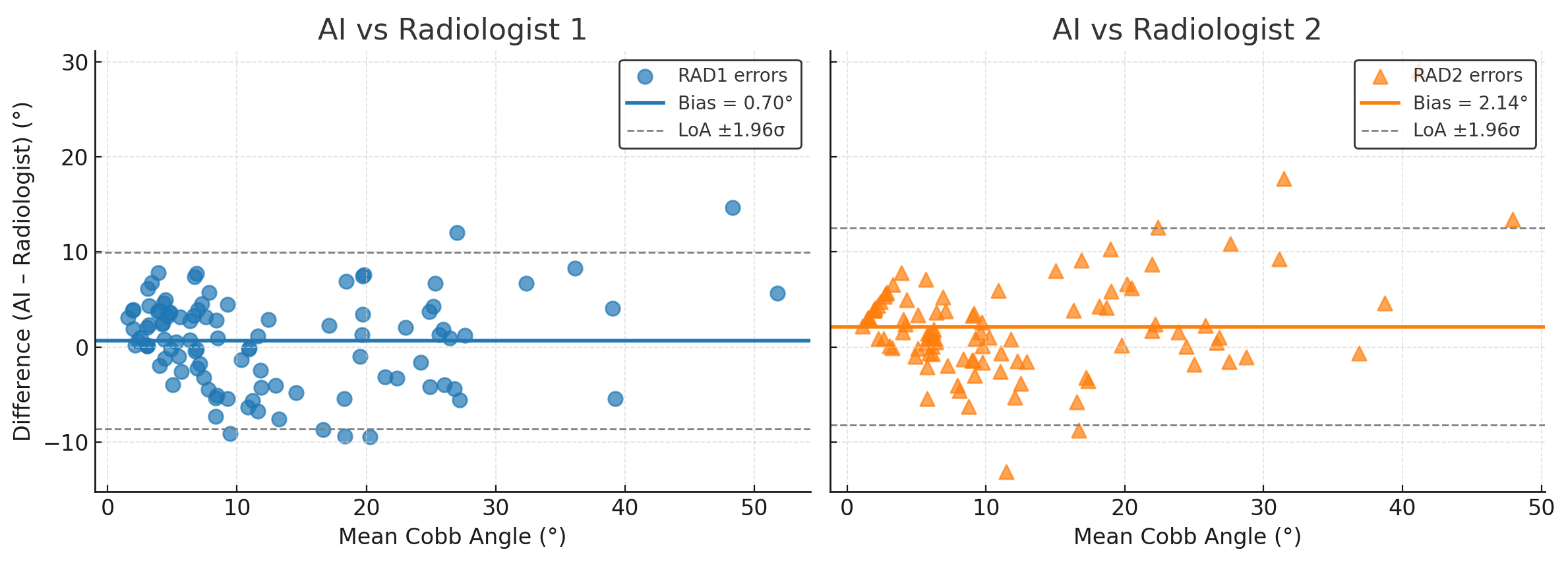}
  \caption{Bland–Altman analysis of Cobb‐angle differences between AI software and each radiologist (RAD 1, RAD 2)}
  \label{fig:bland-altman}
\end{figure}

Bland–Altman analysis comparing AI software with Radiologist 1 yielded a mean bias of $0.70^\circ$ (95 \% CI $-0.23^\circ$ to $1.62^\circ$) and limits of agreement from $-8.59^\circ$ to $9.99^\circ$.  The corresponding mean absolute error was $3.89^\circ$ and the root-mean-squared error was $4.77^\circ$. Versus Radiologist 2, the AI showed a bias of $2.14^\circ$ (95 \% CI $1.10^\circ$ to $3.17^\circ$) with limits of agreement $-8.23^\circ$ to $12.50^\circ$, MAE $3.90^\circ$ and RMSE $5.68^\circ$.  Inter-radiologist comparison produced a bias of $1.44^\circ$ (95 \% CI $0.66^\circ$ to $2.23^\circ$), limits of agreement $-6.44^\circ$ to $9.32^\circ$, MAE $3.30^\circ$ and RMSE $4.25^\circ$. Overall, the AI’s systematic deviation from each radiologist was small and its variability fell within the same range observed between the two human readers (\autoref{tab:primary}, \autoref{fig:bland-altman}).

\begin{table}[!ht]
\centering
\caption{Bland–Altman and error‐metric results.}
\label{tab:primary}
\begin{tabular}{lccccc}
\toprule
Comparison       & Bias (95\% CI)     & LoA (°)            & MAE (°) & RMSE (°) \\
                 & [lower, upper]     & [lower, upper]     &         &          \\
\midrule
AI vs Radiologist 1         & 0.70 [–0.23, 1.62]   & [–8.59, 9.99]   & 3.89   & 4.77   \\
AI vs Radiologist 2         & 2.14 [1.10, 3.17]    & [–8.23, 12.50]  & 3.90   & 5.68   \\
Radiologist 1 vs Radiologist 2 & 1.44 [0.66, 2.23] & [–6.44, 9.32]   & 3.30   & 4.25   \\
\bottomrule
\end{tabular}
\end{table}

Continuous Cobb angles averaged $13.12^\circ \pm 11.10^\circ$ for the AI, $12.43^\circ \pm 10.70^\circ$ for Radiologist 1, and $10.99^\circ \pm 9.39^\circ$ for Radiologist 2 (\autoref{tab:desc}). 

\begin{table}[h]
\centering
\caption{Descriptive statistics of Cobb angles with standard deviation (SD).}
\label{tab:desc}
\begin{tabular}{lcc}
\toprule
Reader           & Cobb angle mean (°) & SD (°) \\
\midrule
AI & 13.12    & 11.10  \\
Radiologist 1    & 12.43    & 10.70  \\
Radiologist 2    & 10.99    & 9.39   \\
\bottomrule
\end{tabular}
\end{table}

\begin{figure}[!ht]
  \centering
  \includegraphics[width=\textwidth]{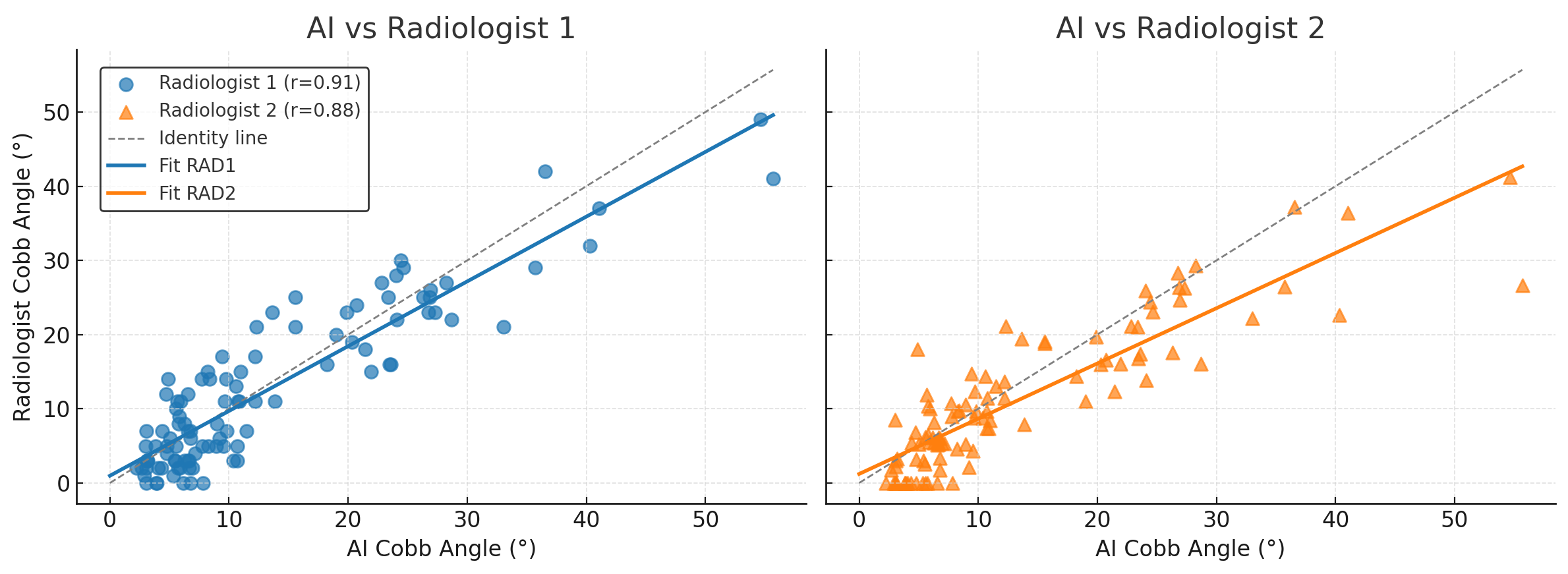}
  \caption{Scatter plots and linear regression of Cobb‐angle measurements for AI software (Carebot AI Bones, Spine Measurement functionality; Carebot s.r.o.) versus each radiologist.}
  \label{fig:regression}
\end{figure}

\newpage
Pearson correlation coefficients were very high for all pairwise comparisons: $r=0.906$ (95 \% CI 0.864–0.936) for AI vs.\ Radiologist 1, $r=0.880$ (95 \% CI 0.827–0.917) for AI vs.\ Radiologist 2, and $r=0.928$ (95 \% CI 0.895–0.951) for Radiologist 1 vs.\ Radiologist 2 (\autoref{tab:corr}, \autoref{fig:regression}).

\begin{table}[h]
\centering
\caption{Pearson’s $r$ with 95\% confidence intervals (CI).}
\label{tab:corr}
\begin{tabular}{lcc}
\toprule
Comparison                   & $r$    & 95\% CI         \\
\midrule
AI vs Radiologist 1          & 0.906  & [0.864, 0.936] \\
AI vs Radiologist 2          & 0.880  & [0.827, 0.917] \\
Radiologist 1 vs Radiologist 2 & 0.928 & [0.895, 0.951] \\
\bottomrule
\end{tabular}
\end{table}

\begin{figure}[!ht]
  \centering
  \includegraphics[width=\textwidth]{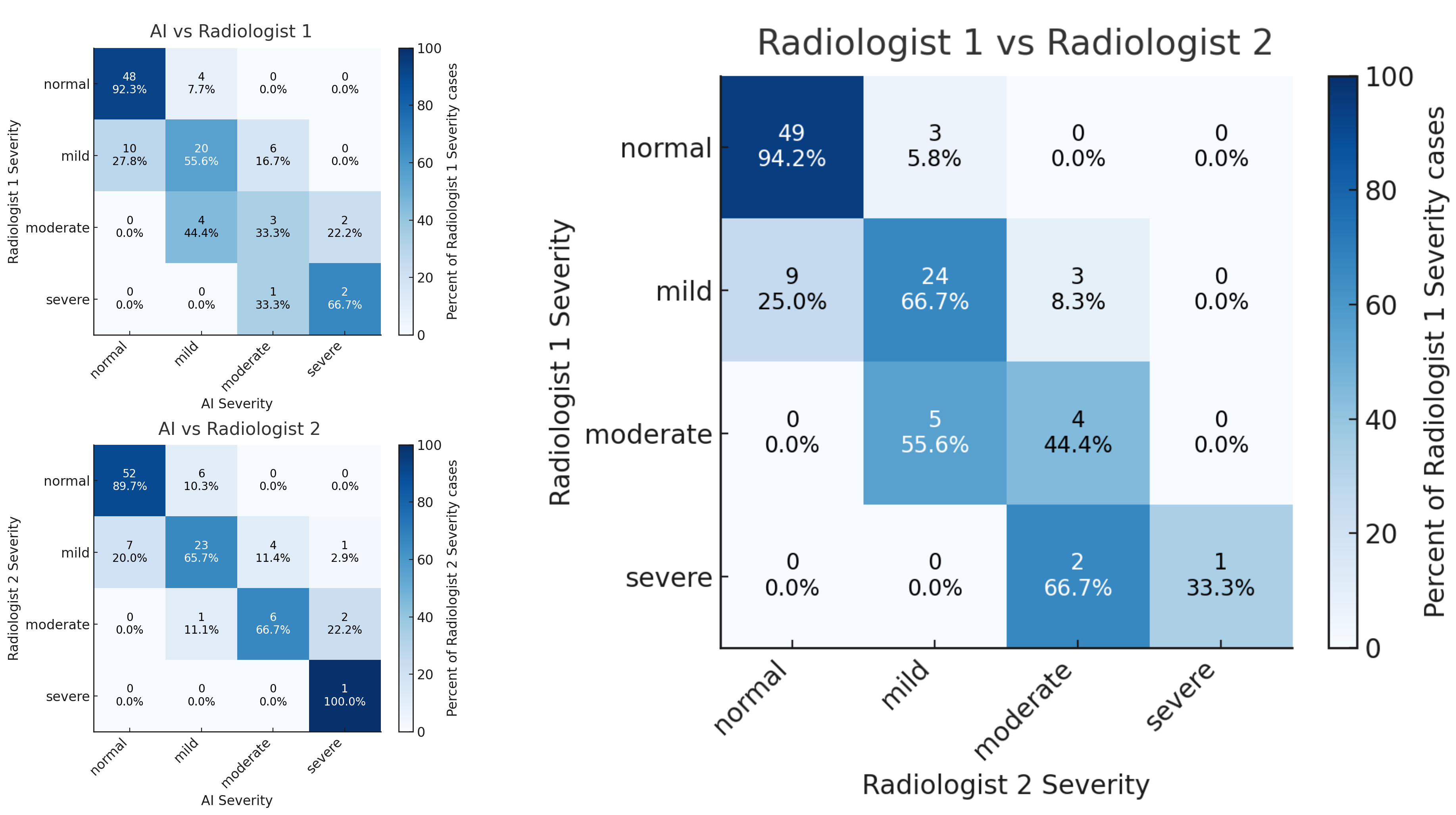}
  \caption{Confusion matrices comparing four-grade scoliosis severity classifications.}
  \label{fig:conf}
\end{figure}

When Cobb angles were classified into four severity grades (none <10°, mild 10–24°, moderate 25–44°, severe $\geq$ 45°), Cohen’s \(\kappa\) indicated moderate to substantial agreement: \(\kappa\)=0.51 for AI vs.\ Radiologist 1, \(\kappa\)=0.64 for AI vs.\ Radiologist 2, and \(\kappa\)=0.59 for Radiologist 1 vs.\ Radiologist 2 (\autoref{tab:kappa}, \autoref{fig:conf}). 

\begin{table}[h]
\centering
\caption{Cohen’s \(\kappa\) for four‐grade severity classification.}
\label{tab:kappa}
\begin{tabular}{lc}
\toprule
Comparison                   & \(\kappa\) \\
\midrule
AI vs Radiologist 1          & 0.51     \\
AI vs Radiologist 2          & 0.64     \\
Radiologist 1 vs Radiologist 2 & 0.59   \\
\bottomrule
\end{tabular}
\end{table}

\newpage

Regarding power analysis, final cohort of 103 whole-spine X-rays exceeds the requirement for MAE precision and lies within the range targeted to bound the 95 \% confidence interval of Cohen’s \(\kappa\) to \(\pm0.15\).

\begin{figure}[!ht]
  \centering
  \includegraphics[width=\textwidth]{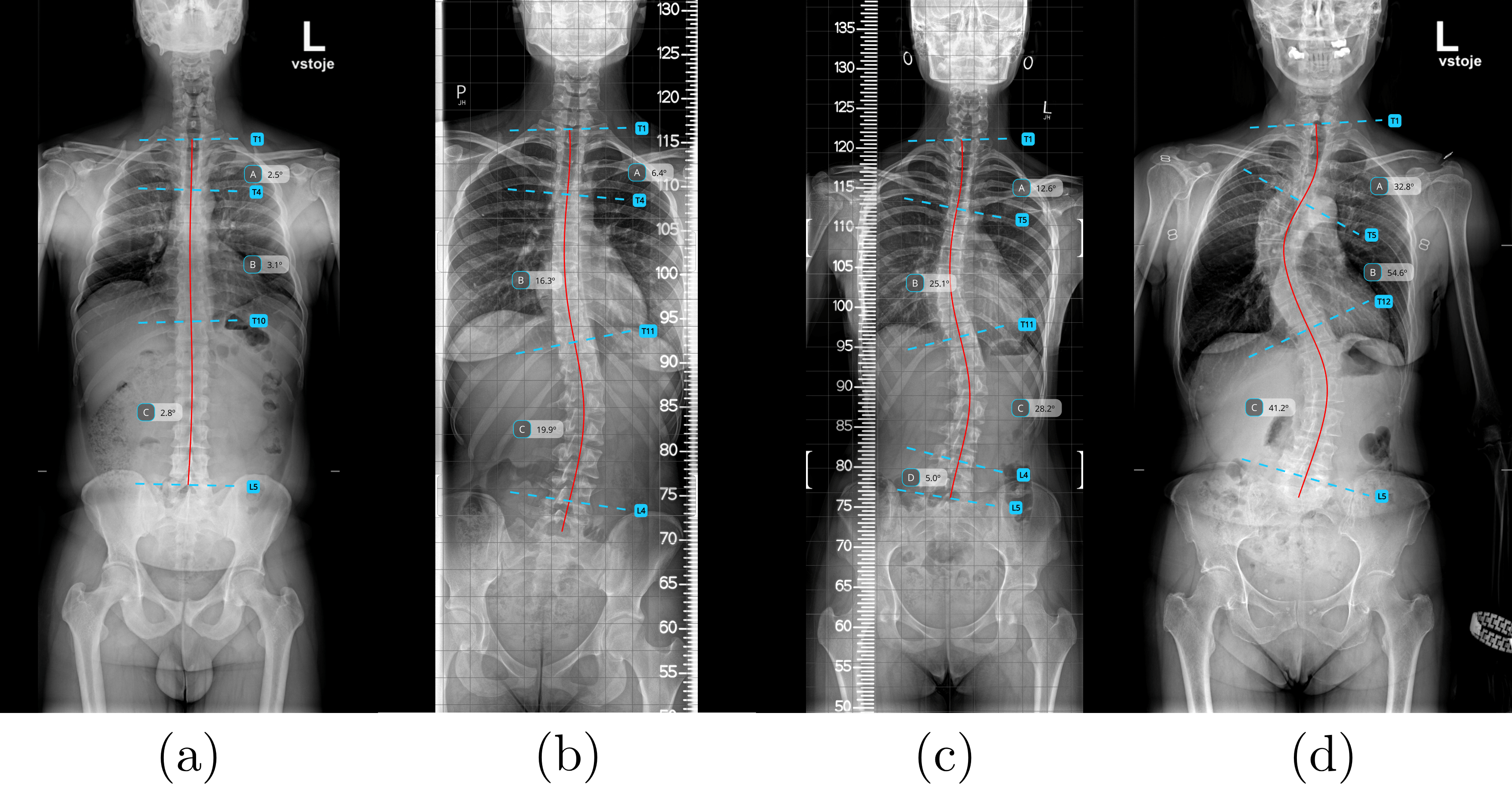}
  \caption{Representative whole‑spine X-rays illustrating four severity classes used in this study, with fully automated overlays generated by the proposed AI software (Carebot AI Bones, Spine Measurement functionality; Carebot s.r.o.).}
  \label{fig:examples}
\end{figure}


\section{Discussion}

In this study we evaluated the performance of AI software (Carebot AI Bones, Spine Measurement functionality; Carebot s.r.o.) for automated Cobb‐angle measurement against two expert radiologists. Across multiple metrics, the AI achieved human-level accuracy. Pearson correlation coefficients exceeded 0.88 for both AI-radiologist comparisons ($r = 0.91$ vs. Radiologist 1; $r = 0.88$ vs. Radiologist 2), rivaling the inter‐rater correlation of $r = 0.93$. Mean absolute error (MAE) between AI and each radiologist ($\approx$3.9°) was virtually identical to that observed between the two radiologists themselves (3.3°), and root‐mean‐squared error (RMSE) differences fell within the same 4–6° range. Bland–Altman analysis revealed small systematic biases (0.7° toward overestimation versus Radiologist 1; 2.1° versus Radiologist 2) and limits of agreement spanning ±8–12°, again mirroring inter‐observer variability (±6–9°). When Cobb angles were categorized into none/mild/moderate/severe grades (\autoref{fig:examples}), Cohen’s \(\kappa\) between AI and each radiologist (\(\kappa\) $= 0.51$ and $0.64$) was on par with the \(\kappa\) $= 0.59$ seen between humans, with most misclassifications confined to adjacent categories.

Recent literature confirms our findings. Landmark‑ or segmentation‑based AI algorithms have achieved MAE values of 2–4° in single‑centre evaluations \cite{horng2019cobb,sun2022comparison}, with ICCs up to 0.994 and Pearson \(r >0.98\). A 2024 meta‑analysis of 17 studies \cite{zhu2025deep} reported a pooled mean error of \(3.0^{\circ}\) (95 \% CI 2.6–3.4°) and confirmed that segmentation approaches outperform landmark detectors (2.4° vs 3.3°). Early multi‑centre efforts have also suggested good generalisability; for example, Hayashi~et al. \cite{hayashi2024deep} reported ICC\,$\approx$\,0.96 across paediatric and adult datasets, while Wong~et al. \cite{wong2023validation} showed mean differences < 3° in a purely paediatric cohort. Nevertheless, most prior studies relied on retrospectively assembled, single‑institution datasets, often restricted to adolescent idiopathic scoliosis and limited to one curve per film, which constrains external validity. By contrast, our multi‑centre design, heterogeneous detector mix, and inclusion of complex, multi‑curve cases demonstrate that expert‑level performance can be maintained under real‑world variability and thus directly address the generalisability gap identified in the current literature.

\subsection{Limitations}

Although our sample was drawn from ten different hospitals, the cohort remained skewed toward paediatric and adolescent examinations (81\% under age 20), which may limit generalizability to older adult populations. A modest number of severe curves ($\geq$ 40°) also reduced precision in the highest‐angle bins. We did not assess the impact of AI integration on actual reporting time or downstream clinical decisions. Finally, while radiologist evaluation served as our comparison reference standard, Cobb angle measurement inherently involves some human error.

\section{Conclusions}

The AI software demonstrated expert-level performance for automated Cobb angle measurement and severity grading, with agreement comparable to that between musculoskeletal radiologists. Its integration into clinical workflows can streamline scoliosis assessment, enhance consistency, and support efficient triage. Further prospective multicenter validation is needed to ascertain its impact on clinical decision-making and workflow efficiency.

%
%
%
%

\end{document}